# Toward Fully Automated Robotic Platform for Remote Auscultation

Ryosuke Tsumura, *Member, IEEE*, Yoshihiko Koseki, Naotaka Nitta, *Member, IEEE*, Kiyoshi Yoshinaka

*Abstract*—Since most developed countries are facing an increase in the number of patients per healthcare worker due to a declining birth rate and an aging population, relatively simple and safe diagnosis tasks may need to be performed using robotics and automation technologies, without specialists and hospitals. This study presents an automated robotic platform for remote auscultation, which is a highly cost-effective screening tool for detecting abnormal clinical signs. The developed robotic platform is composed of a 6-degree-of-freedom cooperative robotic arm, light detection and ranging (LiDAR) camera, and a spring-based mechanism holding an electric stethoscope. The platform enables autonomous stethoscope positioning based on external body information acquired using the LiDAR camera-based multi-way registration; the platform also ensures safe and flexible contact, maintaining the contact force within a certain range through the passive mechanism. Our preliminary results confirm that the robotic platform enables estimation of the landing positions required for cardiac examinations based on the depth and landmark information of the body surface. It also handles the stethoscope while maintaining the contact force without relying on the push-in displacement by the robotic arm.

## I. INTRODUCTION

Most developed countries, especially Japan, are facing a declining birth rate and an aging population, which has caused a severe shortage of young and productive workers. Notably, an increase in the number of patients per healthcare worker is a critical problem [1]. Hence, it will become difficult to maintain the quality of current healthcare services in the future. Robotics and autonomous technology in the medical field are key to addressing this issue. We believe that these technologies, relatively simple and safe diagnosis tasks must be performed without specialists and hospitals.

Since the 1800s, auscultation has been an essential component of clinical examination and is a highly cost-effective screening tool to detect abnormal clinical signs [2]. Auscultation continues to play an important role in the 2020s as cardiopulmonary disease is an important underlying or a direct cause of mortality and morbidity that has a significant impact on quality of life as well as healthcare costs [3]. Additionally, recent studies have reported that auscultation is a potential diagnostic tool for COVID-19 patients and can be used as a follow-up tool for noncritical COVID-19 patients [4], [5]. Since conventional auscultation requires physicians to have physical contact with the patients, remote auscultation is beneficial in terms of protection from infection. Therefore, auscultation is a potential application that is worth being automated with robotic systems that do not rely on physicians.

Recently, electronic stethoscopes have been promising options for addressing the issues of remote auscultations. The electronic stethoscope enables the visualization of sonograms of the heart and lung during auscultation, making it easier to differentiate between several types of heart and lung sounds [2] and provides computer-aided diagnosis such as in coronary artery disease [6]. A difficulty of auscultation is that the efficacy relies significantly on physicians' hearing skills and knowledge. To reduce the subjectivity of auscultation, there is a trend of applying artificial intelligence to stethoscopes [7]–[9]. Although the auscultation's efficacy has been improved with technologies, the auscultation procedure still requires the handheld stethoscopes to be placed on the patient's body by physicians or healthcare workers, which forces them to be physically on site. Additionally, during auscultation, it is necessary for them to place the stethoscope with optimal contact forces against the patient's skin, i.e., to satisfy the patients' safety and eliminate external noise during auscultation. Given that the shortage of healthcare workers will become more serious in the future, especially in rural areas where the medical resources are insufficient, there is a demand for the robotic platform to enable autonomous auscultation with high accuracy and repeatability, not relying on the physician's operation.

Robot-assisted diagnosis has been investigated worldwide, especially in ultrasonography [10], [11]. The types of robotic ultrasonography can be classified as remote operation, human-robot cooperation, and autonomous operation [11], [12]. Particularly, there have been numerous studies on the development of tele-operated robotic ultra-sonography systems [13]–[21]. For fully autonomous robotic ultrasonography, several studies proposed an optimization of the scan trajectory based on the visual servo, real-time 3D reconstruction with the scanned images, and reproduction of the expert's operation in those days [22]–[28]. However, to the best of our knowledge, only a few studies have focused on robot-assisted auscultation. Based on a survey of the relevant scientific literature, we believe that only one paper has reported a tele-operative 3-degree-of-freedom (DOF) robotic system for remote auscultation. However, the proposed system configuration was based on remote control and did not focus on autonomous auscultation [29]. For fully autonomous auscultation, it is ideal that the landing points to place the stethoscope by the robotic system can be determined without operator participation. Additionally, when placing the stethoscope on the landing points using the robotic system, it is necessary to apply an optimal contact force for the patients' safety. For instance, for safety with robotic ultrasonography, the conventional approach is a force-compliance control that controls the position of the robotic arm grasping the ultrasound probe to maintain the contact force within a certain range [24], [27], [28], [30]–[32]. The potential concern for the force-compliance control approach with the robotic arm is that its safety depends on the reliability of the sensor and control algorithm. We assume that the position control should be separated from the contact force control in terms of the reliability of the entire robotic system. Hence, the development of an end-effector that can adjust the contact force safely is necessary. Several unique end-effectors have been proposed that enable the maintenance of the contact force within a certain range regardless of the displacement of the ultrasound probe [21], [33], [34]. The flexible and safe contact by incorporating these end-effectors with the robotic arm may also result in a reliable approach for placing the stethoscope on the landing points autonomously.

*Research supported by AIST internal fund.
R. Tsumura, Y. Koseki, N. Nitta and K. Yoshinaka are with Health and Medical Research Institute, National Institute of Advanced Industrial Science and Technology, Tsukuba, Ibaraki, 305-8564 Japan (e-mail: ryosuke.tsumura@aist.go.jp).

In this study, we aim to develop a robotic auscultation platform that enables estimation of the landing positions and safe placement of the stethoscope at the estimated position. The proposed estimation method is based on registration with a light detection and ranging (LiDAR) camera. For fully autonomous auscultation, the robotic system must recognize the landing positions based on the whole-body shape and place the stethoscope at the determined landing positions, considering the positional relationship between the body and the stethoscope. The LiDAR camera is suitable for acquiring the contour of the entire body surface. Additionally, the emitted laser from the LiDAR is a Class 1 laser that is safe under any conditions of normal use. In this study, the entire body surface was reconstructed by moving the LiDAR camera around the body and combining the sequentially acquired pieces of depth information as point cloud data. With the reconstructed body shape information, the landing positions were estimated based on the anatomical features of the body surface. To place the stethoscope safely on the body surface, an end-effector capable of applying a constant contact force against the body surface with a unique passive mechanism was developed. The passive mechanism comprises a linear spring, a linear servo actuator, and an optical distance sensor and can generate different constant contact forces when the amount of compression of the linear spring is controlled within a certain range in real time.

The contribution of this paper is to establish a proof-of-concept of the robotic platform that enables autonomous positioning of the stethoscope based on external body information while satisfying the patient's safety in terms of the contact between the stethoscope and body surface. To the best of our knowledge, this is the first dedicated robotic system designed for autonomous auscultation. The remainder of this paper is organized as follows. Section II describes the landing position estimation method, including the body-tool registration process, landing position determination, and passive mechanism to place the stethoscope safely. Section III reports the experimental results with an abdominal phantom and five healthy subjects, and a discussion of the proposed

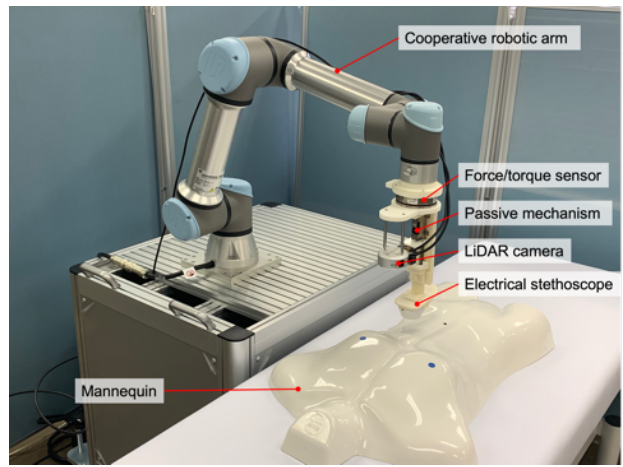

Fig. 1 System overview of robotic auscultation platform.

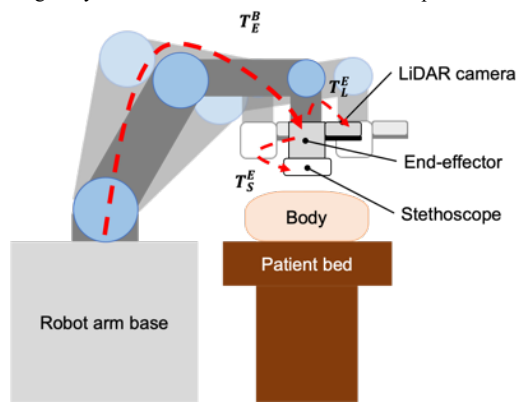

Fig. 2 Data acquisition with LiDAR camera mounted on robotic arm.

method and experimental results. Section IV provides the conclusion and future directions.

## II. METHOD

### A. System Design

The developed system (**Fig. 1**) consisted of the following parts: a medical electronic stethoscope system (JPES-01, MITORIKA, Japan), a 6-DOF cooperative robotic arm (UR5e, Universal Robot, Denmark) for positioning the stethoscope at the determined landing positions, a LiDAR camera (Intel RealSense L515, Intel, USA) to acquire the 3D contour of the body surface as point cloud data, and a WorkStation (Dell Precision 5380, Dell, USA) to synchronize the robotic arm while acquiring the point cloud data. A spring-based passive mechanism is implemented into the end-effector of the robotic arm for gripping the stethoscope adaptively against the tissue surface in terms of scan safety (see Sec II-D). At the base of the end-effector, a 6-axis force/torque sensor (Axia-80-M20, ATI Industrial Automation, USA) was attached to measure the contact force when placing the stethoscope on the body surface. The WorkStation and the server of the robot arm were directly connected to a network (1 GB/s), and the protocol for data transmission was TCP/IP.

The LiDAR camera acquires the depth and color information of the body shape as point cloud data for the registration of the coordinates between the robotic arm and body. Because the LiDAR camera is attached to the end-effector of the robotic arm, the positional relationship between the camera and the robotic arm is kinematically fixed, even when the robotic arm moves around the body. Thus, the position of the acquired point cloud data is linked to the coordinates of the robotic arm. **Fig. 2** illustrates the chain of transformations in the proposed system, composed of the robotic arm, LiDAR camera, and stethoscope. The computed scanning points in the LiDAR camera coordinate space ($P_L$) can be transferred to the coordinate space of the robotic arm's base ($P_B$), and the contact points in the stethoscope coordinate space ($P_S$) can be transferred to the coordinate space of the LiDAR camera ($P_L$), through the following chain of transformations:

$$P_B = T_E^B T_L^E P_L$$
$$P_S = T_S^{E^{-1}} T_L^E P_L \qquad (1)$$

where $T_E^B$, $T_S^E$, and $T_L^S$ represent the base-to-end-effector transformation of the robotic arm, end-effector-to-stethoscope transformation, and stethoscope-to-LiDAR camera transformation, respectively. $T_E^B$ is determined in the robotic arm's controller. $T_S^E$ and $T_L^S$ are calculated for each kinematic relationship using the CAD model.

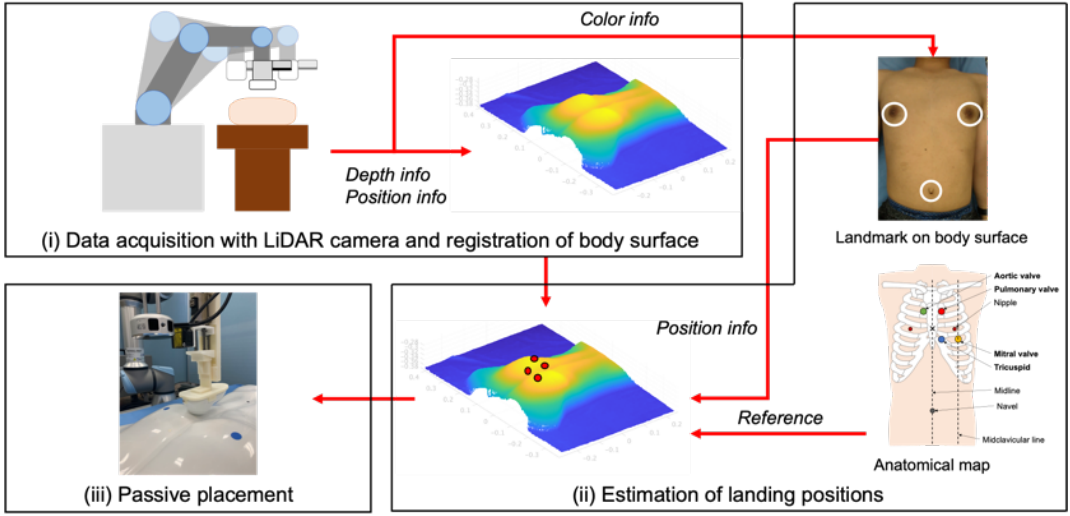

Fig. 3 Pipeline of the estimation of landing positions to place the stethoscope with the developed robotic auscultation platform.

The robotic arm system can be controlled using URScript, which is a unique programming language used in the robotic system. The client PC sends the commands of the URScript via socket communication to the server. The point cloud data was acquired using an Intel RealSense SDK 2.0. A custom-designed software system based on Python programming in Visual Studio Code can synchronize the robotic arm control with the point cloud data reading.

The pipeline for estimating the landing positions to place the stethoscope with the developed robotic auscultation platform is organized into three components: (i) acquisition of the point cloud data for covering the entire chest and registration of the acquired point cloud data to reconstruct the entire chest shape; (ii) estimation of the landing positions based on the reconstructed body shape and the anatomical landmarks on the body surface; (iii) placement of the stethoscope at the estimated positions while maintaining a certain safe contact force, as shown in **Fig. 3**.

### B. Registration of Point Cloud Data

To reconstruct the entire 3D body shape precisely, several datasets of the point cloud under varying LiDAR camera positions must be obtained and registered. The registration of a 3D surface is challenging because noisy data and partially overlapped data must be addressed. The common approach is to combine sampling-based coarse alignment and iterative local refinement, such as iterative closest point (ICP) [35]. As an advanced registration algorithm, pairwise global registration, which is more than an order of magnitude faster than the common registration pipeline and is much more robust to noise, is widely utilized [36]. This approach can also be applied to align multiple surfaces to obtain a model of a large scene or object; this procedure is known was multi-way registration. To further improve the accuracy of registration, it is crucial to obtain the relative position of each dataset. Given that the point cloud data paired with the position data of capturing the point cloud is known, the registration error should be minimized because the system coordinates of the captured datasets can be transformed into global coordinates. As an advantage of the developed system, the LiDAR camera position where each dataset of the point cloud is captured can be obtained accurately based on the encoder implemented in each of the joints of the robotic arm.

Following the proposed multi-way registration, given a set of captured point cloud data of the body surfaces $\{\mathbf{P}_i\}$, it is necessary to estimate a set of poses $\mathbb{T} = \{\mathbf{T}_i\}$ that aligns the surfaces in a global coordinate frame [36]. As each of the captured point cloud data is on the LiDAR camera coordinate space, its coordinates can be transformed to the coordinate space of the robotic arm's base using Eq. (1). The objective function for the multi-way setting can be described as follows [36]:

$$E(\mathbb{T}, \mathbb{L}) = \lambda \sum_i \sum_{(\mathbf{p},\mathbf{q}) \in \mathcal{K}_i} \|\mathbf{T}_i \mathbf{p} - \mathbf{T}_{i+1}\mathbf{q}\|^2$$
$$+ \sum_{i<j} \left( \sum_{(\mathbf{p},\mathbf{q}) \in \mathcal{K}_{i,j}} l_{\mathbf{p},\mathbf{q}} \|\mathbf{T}_i \mathbf{p} - \mathbf{T}_{i+1}\mathbf{q}\|^2 + \sum_{(\mathbf{p},\mathbf{q}) \in \mathcal{K}_{i,j}} \Psi(l_{\mathbf{p},\mathbf{q}}) \right) \quad (2)$$

where $\mathcal{K}_{ij}$ represents a set of candidate correspondences for each pair of surfaces $(\mathbf{P}_i, \mathbf{Q}_i)$. Here, $\mathbb{L} = \{l_{\mathbf{p},\mathbf{q}}\}$ is a line process for the Black-Rangranjan duality [37]. $\Psi(l_{\mathbf{p},\mathbf{q}})$ was set as follows:

$$\Psi(l_{\mathbf{p},\mathbf{q}}) = \mu \left( \sqrt{l_{\mathbf{p},\mathbf{q}}} - 1 \right)^2 \quad (3)$$

To solve the minimization problem, $E(\mathbb{T}, \mathbb{L})$ is first minimized with $\mathbb{L}$. For $E(\mathbb{T}, \mathbb{L})$ to be minimized, the partial derivative with respect to each $l_{\mathbf{p},\mathbf{q}}$ is calculated as follows:

$$\frac{\partial E}{\partial l_{\mathbf{p},\mathbf{q}}} = \|\mathbf{p} - \mathbf{T}\mathbf{q}\|^2 + \mu \frac{\sqrt{l_{\mathbf{p},\mathbf{q}}} - 1}{\sqrt{l_{\mathbf{p},\mathbf{q}}}} = 0 \quad (4)$$

Solving for $l_{\mathbf{p},\mathbf{q}}$ with Eq. (2),

$$l_{\mathbf{p},\mathbf{q}} = \left( \frac{\mu}{\mu + \|\mathbf{p} - \mathbf{T}\mathbf{q}\|^2} \right)^2 \quad (5)$$

Then, $E(\mathbb{T}, \mathbb{L})$ is minimized with respect to all poses $\mathbb{T}$. $\mathbf{T}_i^k$ is denoted as the $i$-th transformation pose calculated in the previous iteration. $\mathbf{T}_i$ can be linearized with a 6-vector $\xi_i = (\alpha_i, \beta_i, \gamma_i, a_i, b_i, c_i)$ as follows:

$$\mathbf{T}_i \approx \begin{pmatrix} 1 & -\gamma_i & \beta_i & a_i \\ \gamma_i & 1 & -\alpha_i & b_i \\ -\beta_i & \alpha_i & 1 & c_i \\ 0 & 0 & 0 & 1 \end{pmatrix} \mathbf{T}_i^k \quad (6)$$

When the $6|\mathbb{T}|$-vector that collates $\{\xi_i\}$ is $\Xi$, $E(\mathbb{T}, \mathbb{L})$ becomes a least-squares objective on $\Xi$. Then, the objective function can be minimized by solving the linear system as follows while updating $\mathbf{T}_i$:

$$\mathbf{J_r^T J_r} \Xi = -\mathbf{J_r^T r} \quad (7)$$

where $\mathbf{J_r}$ and $\mathbf{r}$ represent the Jacobian matrix and residual vector.

Based on the multi-way registration, the proposed registration pipeline is as follows: First, the LiDAR camera mounted on the end-effector of the robotic arm moves several points away from the body and captures the body surface. Next, each of the coordinates of the captured dataset is transformed to the coordinates of the robotic arm base (global coordinates). With the datasets of the point cloud represented in the global coordinate, multi-way registration is performed. Each of the datasets is combined as one dataset by using the estimated transformation in each of the datasets. Finally, a filter that removes inlier and outlier point clouds is applied to the combined data.

*C. Landing Position Estimation*

Auscultation is mainly performed to examine the circulatory and respiratory systems, namely heart and breath sounds. This study focused on circulatory examination first. For the examination of the circulatory system, four landing positions for placing the stethoscope are used to listen to the sounds of the tricuspid, mitral, pulmonary, and aortic valves. The sounds of the tricuspid, mitral, pulmonary, and aortic valves can be roughly heard to the left of the lower part of the sternum near the 5th intercostal space, over the apex of the heart in the left 5th intercostal space at the midclavicular line (approximately 10 cm from the midline), over the medial end of the left 2nd intercostal space, and over the medial end of the right 2nd intercostal space, respectively [38][1]. The robotic system must recognize each of the positions to hear the sounds of the corresponding valves autonomously.

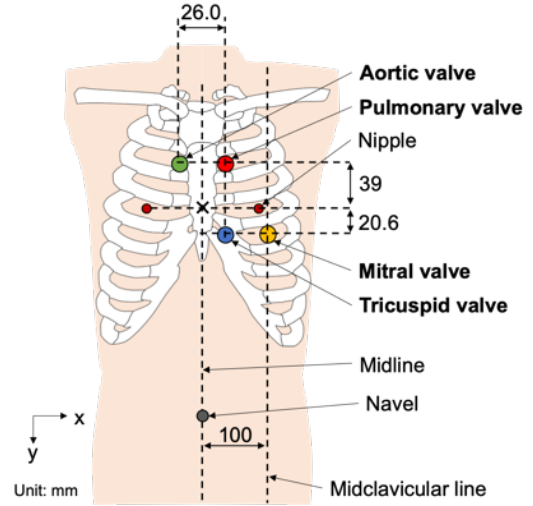

Fig. 4 Anatomical map showing the location of each valve on the body surface to place the stethoscope.

To estimate each position on the body surface, the nipple and navel may be applicable landmarks that the robotic system can recognize easily on the body surface. The nipple is mostly located on the 4th intercostal space [39]. Its location can be the landmark to find each landing position based on the anatomical relationship between each rib and its intercostal spaces. In [40], the height of the intercostal space and ribs were measured using CT scan data. Based on the statistical data, the length along the cephalocaudal direction between the 4th intercostal space (nipple's position) and 2nd intercostal space (aortic and pulmonary valves' position) was 39 mm, and the length along the cephalocaudal direction between the 4th intercostal space (nipple's position) and 5th intercostal space (mitral and tricuspid valves' position) was 20.6 mm on average. These lengths can be utilized as a reference for the valves' height on the robotic system coordinate ($y$-axis). The position of the midline on the abdomen is necessary as a reference to estimate the width of the valves on the robotic system coordinate ($x$-axis). The midline can be identified by connecting the center position of the nipples to the navel's position. The midclavicular line where the mitral valve is on is approximately 100 mm away from the midline along the x-axis [38]. Because other valves are located near the medial end of the intercostal space, the stethoscope should be placed around the edge of the sternum. The width of the male sternum is 25.99 mm on average [41]. Then, we assume that the positions of the aortic and pulmonary valves along the horizontal axis are ± 13 mm with reference to the midline. Based on the references of each valve's position, an anatomical map depicting each landing position on the body surface in 2D space was created, as shown in **Fig. 4**.

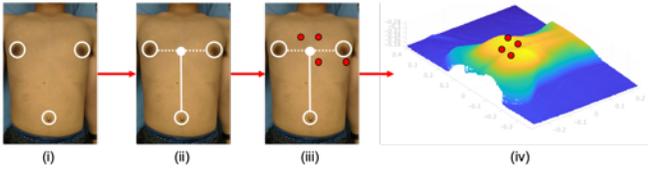

Fig. 5 Procedure to estimate the landing positions based on landmarks on body surface and anatomical map.

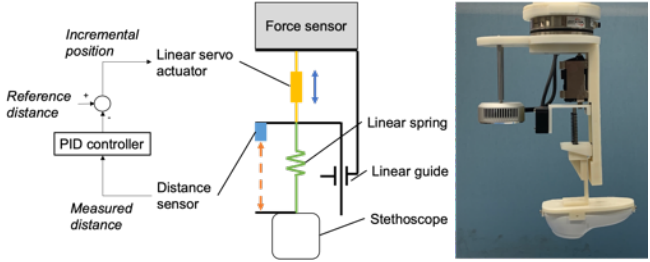

Fig. 6 Configuration of the passive mechanism to maintain the variable contact force within a certain range.

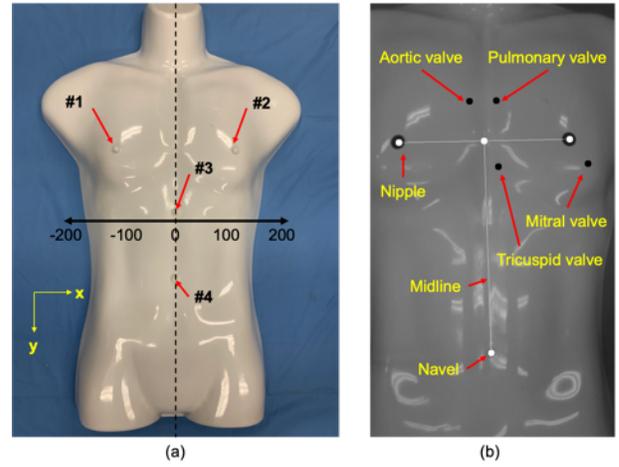

Fig. 7 Target locations on the mannequin in (a) the multiway-registration experiment and (b) the estimation experiment.

In short, to estimate each landing position to place the stethoscope, it is necessary to identify the locations of the nipples and navel from external body information. By combining the identified locations of the nipples and navel with the anatomical map on the body surface, each landing position for hearing the sounds of the four valves can be estimated. The entire procedure for estimating the landing positions is as follows (**Fig. 5**): (i) First, the locations of the nipples and navel are extracted from the color image as 2D pixel locations using a template matching method. (ii) The direction of the midline on the abdomen was identified by connecting the center point between the extracted locations of the nipples and navel. (iii) With the identified midline location, each landing position on the 2D pixel space of the color image can be estimated based on the aforementioned anatomical map (**Fig. 4**). (iv) The extracted 2D pixel locations in the color image are then projected onto the 3D coordinate space of the reconstructed body surface with the conversion based on the intrinsic camera parameters.

### D. Constant Force Passive Mechanism

To place the stethoscope safely, a unique end-effector with a spring-based passive mechanism was developed. The entire configuration of the developed end-effector is shown in **Fig. 6**. The role of this passive mechanism is to generate a safe, constant contact force for the stethoscope against the body surface, regardless of the push-in displacement by the robotic arm when placing the stethoscope on the estimated landing positions. A certain error will exist between the actual position of the stethoscope and the intended position on the body surface, due to errors in the proposed estimation method or displacements of the body surface (e.g., breathing motion). The end-effector is required to compensate for the error while maintaining the contact force within a certain range. The developed end-effector comprises a linear servo actuator (L12-20PT, MigthyZap, South Korea), a general linear spring, an optical distance sensor (ZX-LD100L, Omron, Japan), and a linear guide (SSE2B6-70, Misumi, Japan). The linear servo actuator moves back and forth while maintaining a constant amount of linear spring compression. The linear spring coefficient was 0.45 N/mm in this study, and two springs were inserted. The amount of compression was measured in real time with the optical distance sensor. The linear servo actuator was controlled via an Arduino-based controller (IR-STS01, MigthyZap, South Korea). The value measured by the optical distance sensor was transferred to the control PC via a DAQ tool (Analog Discovery 2, Digilent, USA). A custom-designed software system based on Python programming in Visual Studio Code synchronizes the control of the linear servo actuator with reading data from the optical distance sensor. A proportional-integral-derivative (PID) control scheme is used to control the position of the linear servo actuator based on the optical distance sensor feedback.

### E. Experimental Setup

To validate the proof-of-concept of the robotic platform that enables autonomous positioning of the stethoscope based on external body information while satisfying the patient's safety, we conducted three main types of experiments. In all experiments, a white male torso mannequin was used as the target of auscultation. Note that as the mannequin does not have the nipples and navel, we set markers at each of the locations on the mannequin.

First, the accuracy of the reconstructed body surface with the RGB-D camera described in Section II-B was evaluated. We acquired the point cloud data at five positions (-200, -100, 0, 100, and 200 mm from the central axis of the body along the x-axis in **Fig. 7**) for multi-way registration. Moreover, we compared the performance of the multi-way registration with and without feedback from the LiDAR camera's position (Eq. (1)). Additionally, because the accuracy of the point cloud depends on the distance between the camera and objects, the point cloud data was acquired by changing the height of the LiDAR camera's position in three steps (250, 300, and 350 mm from the top of the chest). To evaluate the accuracy of the multi-way registration, we moved the robot arm to four targets on the mannequin (**Fig. 7**) based on the reconstructed body surface data and measured the distance between the tip of the end-effector and the targets in 3D coordinate space. The position of each target was manually indicated on the reconstructed body surface data in this experiment.

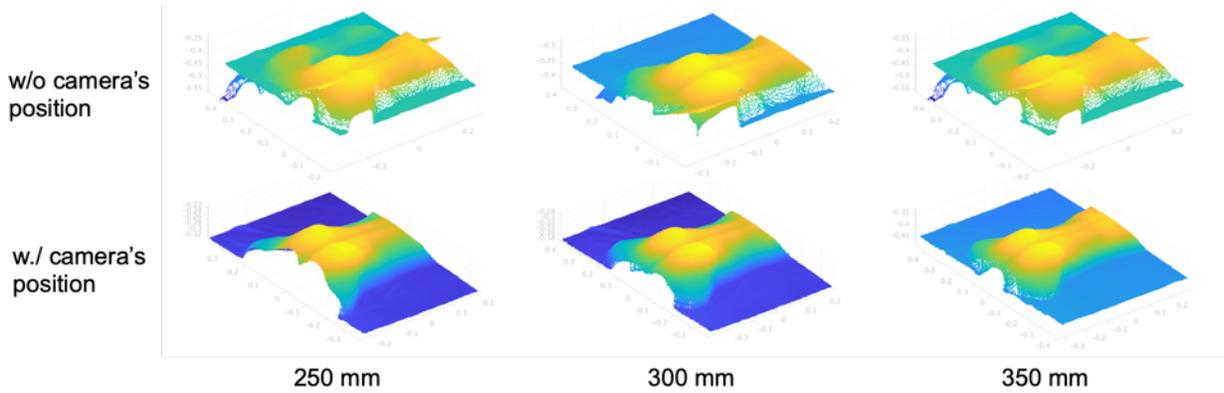

Fig. 8 Results of multi-way registration with and without LiDAR's camera position feedback varying camera height.

Additionally, to eliminate other factors without the registration method (e.g., error due to the assembly of mechanical parts), a jig was attached to the robotic arm to identify the center of the end-effector. Twelve trials were performed for each condition.

Second, the accuracy of the estimation method of the landing positions to place the stethoscope described in Section II-C was evaluated. The reconstructed body surface data used in the previous experiment (LiDAR camera height: 300 mm) was utilized to estimate the four landing positions. We set the markers on each of the targeted landing positions on the mannequin based on the aforementioned anatomical map (see **Fig. 4**) as the ground truth. The position of the mannequin was slightly changed in twelve patterns at random.

Third, the safety of the developed end-effector with the passive mechanism described in Section II-D was evaluated by measuring both the static and dynamic contact forces. The static contact force was measured after the stethoscope was pressed 5 mm against the ground. In this experiment, we set three-way contact forces (5, 10, and 15 N). Twelve trials were performed for each condition. The dynamic contact force was measured when the stethoscope moved from the air to the body surface of the mannequin. The initial position of the stethoscope was 5 mm from the body surface along the z-axis. The stethoscope was moved 10 mm along the z-axis and was thus pressed 5 mm against the body surface. The targeted contact force in this experiment was set to 5 N.

III. RESULTS

A. Reconstruction of Body Surface

**Fig. 8** shows the results of the multi-way registration with and without the feedback of the LiDAR camera's position. Although the reconstruction without the feedback of the LiDAR camera's position was misaligned under all height conditions, the reconstruction with the feedback could show the whole-body shape successfully. Additionally, **Fig. 9** shows the result of the registration error varying with the height of the LiDAR camera. The results indicate that the error decreases depending on the distance between the camera and the target. A two-tailed student's t-test with a 90% confidence interval was used to determine if there were significant differences in the accuracy due to the camera's height. There was a significant difference in the error in the 3D space coordinates for camera heights between 250 mm and 350 mm ($p<0.05$).

B. Estimation of Landing Positions

**Fig. 10** shows the results of the positioning errors at each of the estimated landing positions. The error in the tricuspid valve was small compared to that in the other valves. The two-tailed student's t-test with a 90% confidence interval was also used in the positioning accuracy depending on the target valve's position, and there was a significant difference in the error of the 3D space coordinate between the tricuspid valve and other valves ($p < 0.01$).

C. Contact Force with Passive Mechanism

**Figs. 11 (a)** and **(b)** show the results of the static and dynamic contact forces generated by the developed passive mechanism, respectively. The results of the static force showed that the generated contact forces were precisely achieved to the targeted force under all conditions. Based on the dynamic force results, the measured time-series contact force increased when the surface was touched (0.5 s) and slightly exceeded the targeted contact force; however, the measured force was immediately adjusted to match the targeted force. The maximum measured contact force was 5.36 N which was an error of 7.2% of the targeted force.

IV. DISCUSSION

The results of this study demonstrate the proof-of-concept of the autonomous robotic auscultation platform for circulatory examination. The robotic platform enables the estimation of the landing positions to hear the sounds of four cardiac valves based on the depth information of the body surface and the anatomical map and places the stethoscope while safely maintaining the contact force within a certain range. While the multi-way registration could reconstruct the 3D body surface visually because of the feedback of the LiDAR camera's position, the registration error in the 3D space occurred in the range of 5.1 to 7.6 mm on average. Considering that the resolution of the LiDAR camera used in this study was between 5 and 14 mm, the performance of the registration could be maximized. Additionally, the LiDAR camera's resolution was improved slightly, depending on the distance between the camera and objects. However, because of its nature, the resolution is improved by decreasing the distance, a certain distance between the camera mounted on the robotic arm and the body surface may be required in terms of patient safety during the registration process. Also, the positioning errors to the estimated landing positions were between 9.3 and 17.4 mm on average. These errors included the registration error, estimation error, and mechanical assembly error. In the estimation process, template matching was applied for image processing. Once the detected locations of the nipples and navel are slightly shifted in the image processing, the midline identified based on those nipples and navel' locations is misaligned, which causes the estimation errors of all valves because each of the valves' location is estimated with the relative position from the midline and nipples' locations. Particularly, the error may be increased significantly in the landing position away from the center position of the nipples, which was utilized to identify the base of the midline. Although there are certain errors in the estimation of the landing position, these are acceptable if the position of the stethoscope can be fine-tuned based on the acquired sound, which is similar to the visual servo control.

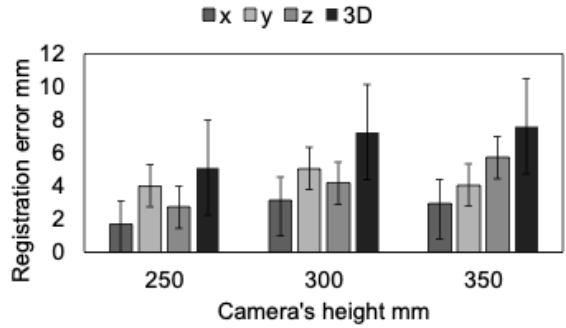

Fig. 9 Results of the multi-way registration error with varying camera height.

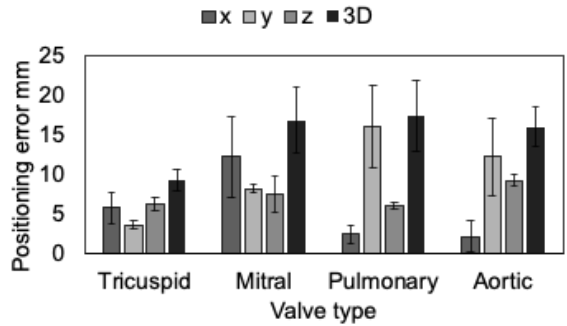

Fig. 10 Results of the positioning error at each of the estimated landing positions on the body surface.

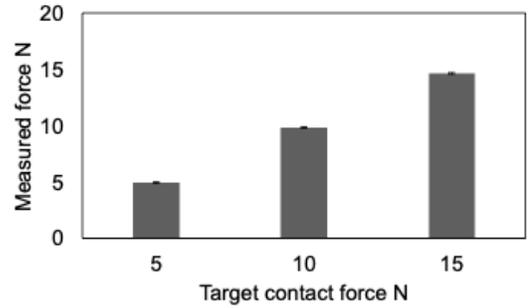

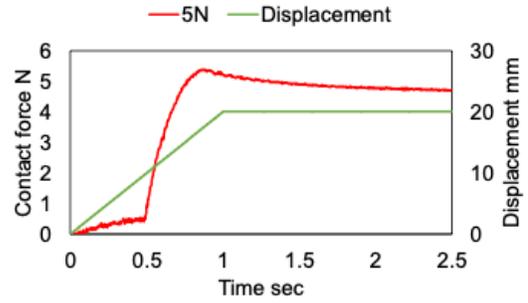

Fig. 11 Results of (a) the static contact force and (b) the dynamic contact force generated by the passive mechanism.

The contact force generated by the passive mechanism could be maintained toward the targeted force within a certain range regardless of the pushing displacement of the stethoscope. The targeted contact force used in this study was tentative because there was no reference regarding the contact force during auscultation via our survey. We assume that the optimal contact force can be determined based on the quality of the acquired sound and the comfort of the patients. While it is necessary to further investigate the optimal contact force, the developed passive mechanism enables the generation of a variable contact force by changing the amount of the spring's compression. The passive mechanism can be applicable if the required contact force varies depending on the valve locations or individual differences, such as sex and the amount of fat. For additional safety, the force sensor attached to the end-effector can be utilized to monitor the generated contact force in real time to detect abnormal status such as sudden patient movement during the procedure.

A significant limitation of the robotic auscultation platform is the lack of validation in human trials. The quality of each cardiac valves' sound cannot be evaluated in the current setup. Since the applicable error in the estimation can be determined based on the quality of the sound, we recognize that it is necessary to perform a comparative study on a larger number of subjects. Such subjects might include those with a variation in body surface area, excess body fat, and female subjects. Additionally, the current setup did not have a respiratory function, which may affect the performance of multi-way registration. Given that the acquired data with the LiDAR camera is slightly shifted by a few millimeters depending on the breathing cycle, it is necessary to investigate whether the multi-way registration can compensate for the displacement of the body surface derived from respiration.

V. CONCLUSION

This manuscript presents a conceptual model of the robotic auscultation platform that enables the autonomous positioning of the stethoscope to satisfy the patient's safety in terms of contact force. Our preliminary results demonstrated that the robotic platform enables estimation of the landing positions to hear the sounds of four cardiac valves based on the depth and landmark information of the body surface and places the stethoscope while maintaining the contact force without relying on its pushing displacement. The developed robotic platform has the potential to address the critical issue of the increase in the number of patients per healthcare worker. The use of this technology may further enhance the efficiency of screening for abnormal clinical signs, including COVID-19.